\documentclass[5p,authoryear]{elsarticle}

\def\tsc#1{\csdef{#1}{\textsc{\lowercase{#1}}\xspace}}
\tsc{WGM}
\tsc{QE}

\newcommand{\myLabel}[2]{\label{#1:#2}#2}
\newcommand{\myRef}[2]{\ref{#1:#2}}
\makeatletter
\DeclareRobustCommand\onedot{\futurelet\@let@token\@onedot}
\def\@onedot{\ifx\@let@token.\else.\null\fi\xspace}

\newcommand{\mySection}[1]{\section{\myLabel{section}{#1}}}

\usepackage{soul}

\usepackage{breqn}

\usepackage{xspace}

\usepackage{hyperref}

\usepackage{svg}

\begin{document}
\let\WriteBookmarks\relax
\def\floatpagepagefraction{1}
\def\textpagefraction{.001}

\title{%
       A New Hope for PRNU on Modern Smartphones? Evidence That Device-Specific Fingerprints Survive Computational Photography\\
}

\let\printorcid\relax
\author[1,2]{Benjamin Loison}%

\ead{benjamin.loison@polytechnique.edu}

\affiliation[1]{organization={Agence Ministérielle pour l'IA de Défense},%
            country={France}}

\author[3]{Quentin Bammey}%

\author[1]{Thibaud Ehret}
\author[2]{Rafael Grompone von Gioi}
\author[2,4]{Pablo Musé}
\author[2]{Marina Gardella}

\affiliation[2]{organization={ENS Paris-Saclay, Université Paris-Saclay, CNRS, Centre Borelli},
            city={Gif-sur-Yvette},
            country={France}}

\affiliation[3]{organization={Télécom Paris, Institut Polytechnique de Paris, LTCI},%
            city={Palaiseau},
            country={France}}
            
\affiliation[4]{organization={IIE, Facultad de Ingenier\'ia, Universidad de la Rep\'ublica},
            country={Uruguay}}

\begin{abstract}
Photo Response Non-Uniformity (PRNU) is widely regarded as the most reliable marker for source camera identification. Recent studies, however, report high correlations between images and fingerprints from different modern smartphones, which calls into question the discriminative power of PRNU on these devices. Using 30 recent smartphones under controlled acquisition conditions (e.g., no digital zoom), we investigate whether the fingerprint itself is at fault or the way it is compared. We find that a component shared by all fingerprints of the same model inflates the standard signed peak-to-correlation energy (sPCE). To counter this, we propose $\Delta$sPCE: the sPCE between a query image and a candidate camera’s fingerprint, minus the mean sPCE between that image and the fingerprints of the remaining candidates, in the spirit of T-norm score normalization. $\Delta$sPCE improves attribution accuracy, suggesting that the device-specific PRNU signal is still present. 
Although the limited dataset size and the dependence of $\Delta$sPCE on the candidate camera set currently restrict operational applicability, these results suggest that PRNU-based attribution of modern smartphones remains feasible when fingerprints are compared appropriately.
\end{abstract}

\maketitle

\mySection{Introduction}

This work focuses on the Photo Response Non-Uniformity (PRNU), a deterministic type of image noise caused by minor manufacturing imperfections in the image sensor. These imperfections cause each individual pixel to have a small, consistent multiplicative gain error, leading it to slightly overestimate or underestimate the intensity of incoming light. These gain variations are unique to each pixel, remain constant over time and across images, and serve as a distinct fingerprint of the sensor that captured the image~\citep{largescale_goljan, Lukas_2006}. PRNU-based methods meet multiple forensic needs; for example, they can be used to assign an image to its source camera or to determine whether two images were taken with the same camera.

Although PRNU has been studied for more than two decades~\citep{Lukas_2006}, cameras and their image processing pipelines have evolved substantially during this time~\citep{delbracio2021mobile}. Modern cameras, particularly smartphones, rely on increasingly sophisticated pipelines, with manufacturers applying proprietary processing steps that can alter or suppress sensor-specific characteristics. These changes have been reported to affect the reliability and discriminability of PRNU-based methods for camera source identification. For instance,~\cite{9393356} and~\cite{9413611} showed that advanced image processing techniques used in smartphones affect the extraction and use of PRNU.

\newcommand{\ImageSourceIdentification}{Standard PRNU-based source identification}
\newcommand{\sPCEbasedPRNUDetection}{PRNU detection}

Motivated by the findings of these works, we revisit PRNU-based source camera identification within modern smartphone imaging pipelines from two complementary perspectives. First, we focus only on images processed using the device's default processing pipeline. This provides a controlled setting for assessing whether PRNU-based source identification remains possible when images from the same device undergo consistent processing. Second, we propose a
background model, specific to each image, smartphone model, and software version,
that uses the PRNU estimates from other cameras of the same smartphone model and software version as a reference, rather than relying solely on the candidate camera's PRNU estimate.

Our overall contributions are twofold. First, we evaluate classical PRNU-based source camera attribution on 30 recent smartphones from three major manufacturers, using images with comparable processing pipelines, and confirm the limitations of the conventional sPCE score.
We then introduce $\Delta\text{sPCE}$, a simple score normalization method that leverages fingerprints from other devices that share the same model and software version. Our experiments show substantial improvements in threshold-based attribution, particularly for Apple smartphones. These findings suggest that device-discriminative information remains detectable despite the modern computational photography pipelines.

\newcommand{\ImageSourceIdentificationPRNUDetection}{PRNU detection}
\newcommand{\ImageSourceIdentificationPRNUDetectionDifferenceOfsPCEsBasedPRNUDetection}{Difference of sPCEs-based PRNU detection}
\newcommand{\ImageSourceIdentificationPRNUDetectionDifferenceOfsPCEsBasedPRNUDetectionLabel}{\ImageSourceIdentification:\ImageSourceIdentificationPRNUDetection:\ImageSourceIdentificationPRNUDetectionDifferenceOfsPCEsBasedPRNUDetection}

\newcommand{\Experiments}{Experiments}
\newcommand{\RelatedWorks}{Related works}

The remainder of this article is organized as follows.
Section~\myRef{section}{\RelatedWorks} reviews prior work on PRNU-based source camera identification in modern smartphones.
Section~\myRef{section}{\ImageSourceIdentification} describes how a camera's PRNU fingerprint is estimated and how its presence in a query image is detected.
Section~\myRef{section}{\ImageSourceIdentificationPRNUDetectionDifferenceOfsPCEsBasedPRNUDetectionLabel} introduces the proposed $\Delta$sPCE metric, which is computed against a background model built from smartphones sharing the same software version.
Section~\myRef{section}{Dataset} presents an image dataset captured with the default processing pipelines of recent smartphones.
Section~\myRef{section}{\Experiments} uses this dataset to evaluate PRNU estimation and source identification under those pipelines, and to assess the benefit of the proposed background model.
Section~\myRef{section}{Discussion} discusses the implications and limitations of our findings, and why they constitute a new hope for source camera identification on modern smartphones.
Finally, Section~\myRef{section}{Conclusion} summarizes the conclusions of this work.

\mySection{\RelatedWorks}

\cite{11346806,vázquezpadín2025diagonalartifactssamsungimages,10.1145/3785353.3815090} and \cite{10.1145/3658664.3659650,10715103} have analyzed different sources of Non-Unique Artifacts (NUAs) highlighted by \cite{10.1007/978-3-030-69449-4_15}, \cite{9393356} and \cite{9413611} on respectively Apple iPhones and Samsung Galaxy smartphones and Adobe software. \cite{10.1145/3658664.3659654} also considered visible watermarks applied to Xiaomi Mi 9 and artifacts in the Discrete Fourier Transform magnitude of Samsung A50 images. The characterization of such an unwanted common signal across camera instances enables the detection of potentially smartphone models, images, and parts of images suffering from NUAs, allowing their exclusion from forensic practitioner analysis~\citep{9897923} or, if possible, their removal.
However, such removal of NUAs requires extensive analysis for each image processing pipeline and may be incomplete. The background model improvement we introduce does not require a prior-specific image-processing pipeline analysis.

\cite{Chen_2008} introduce the zero-mean noise residual post-processing to reduce artifacts introduced by color interpolation and row/column-wise processing in imaging sensors and circuits. %

While the sPCE introduced by \cite{Goljan2008} normalizes the squared correlation at zero shift by the average squared correlation away from zero shift (see Section~\myRef{section:\ImageSourceIdentification}{\sPCEbasedPRNUDetection}), it does not leverage fingerprints from other cameras that share the same image processing pipeline.
In speaker verification, test normalization (T-norm) normalizes a similarity score by subtracting the mean and dividing by the standard deviation of scores obtained for the same test sample against a background set of speaker models~\citep{10.1006/dspr.1999.0360}.
Inspired by this approach, we introduce the $\Delta\text{sPCE}$, which subtracts from each candidate's sPCE the mean sPCE obtained with fingerprints from other cameras that share the same model and software version. To our knowledge, this application of score normalization is novel in PRNU-based source camera attribution.
Unlike full T-norm, $\Delta\text{sPCE}$ omits the division by the standard deviation, which we found to reduce attribution accuracy in preliminary experiments.

In recent years, data-driven approaches leveraging neural networks have shown some improvement in camera fingerprint estimation and detection~\citep{8713484,LI2026132285,Cozzolino2020}. Nevertheless, deep learning methods have limitations, notably in training set scarcity, explainability, and robustness. %

\mySection{\ImageSourceIdentification}

PRNU-based image source camera identification involves two main steps: estimating the camera's PRNU fingerprint and detecting this fingerprint in a query image. Several approaches have been proposed for both steps, and recent work has provided comprehensive evaluations of their performance~\citep{52678902fc7f478ebeb35078e7a5e591}. %
In this section, we describe the operators used in this work, which follow those adopted in previous studies on PRNU-based source identification, particularly those of~\cite{9393356}, to ensure consistency with existing experimental protocols and to enable meaningful comparison with their results. The following subsections describe these operators and the corresponding source attribution procedure.

\subsection{PRNU estimation}

Throughout this article, bold symbols denote matrices while scalar quantities appear in regular font. All operations are understood to be element-wise.

The camera PRNU $\textbf{K}$ can be estimated using the noise residuals $\textbf{R}_1, \ldots, \textbf{R}_L$ extracted from images $\textbf{I}_1, \ldots, \textbf{I}_L$ taken by the same camera using a denoiser $\mathcal{D}$, as
\begin{equation}
\textbf{D}_i = \mathcal{D}(\textbf{I}_i),
\qquad
\textbf{R}_i = \textbf{I}_i - \textbf{D}_i,
\end{equation}
where $\textbf{D}_i$ is the denoised version of $\textbf{I}_i$.
In this work, we use the wavelet denoiser described in~\citep{10.1109/ICASSP.1999.757535}.
The maximum likelihood estimator~\citep{Chen_2008} is then used to combine the noise residuals and estimate the camera's PRNU:
\begin{equation}\label{eq:mle}
    \hat{\textbf{K}}_{\text{MLE}} = \frac{\sum_{l=1}^L \textbf{R}_l \textbf{D}_l}{\sum_{l=1}^L \textbf{D}_l^2}.
\end{equation}

To further refine the distinguishability of the estimated PRNU, the zero-mean post-processing introduced by~\cite{Chen_2008} is applied. Specifically, for each pixel, the mean value of its column is subtracted, followed by the subtraction of the mean value of its row (or vice versa).
The purpose of this operation is to reduce artifacts introduced by color interpolation and row/column-wise processing in imaging sensors and circuits.

\begin{table*}[t]
\begin{tabular}{l|c|c|c|c|c|c}
Smartphone-software-version & Inst. & Ref. img. & Ref. PRNU & Test img. & Match corr. & Mismatch corr. \\
\hline
Google Pixel 10 Pro XL              &  4 &  80 &  4 &  80 &  80 &  240 \\
Google Pixel 10 Pro                 &  6 & 120 &  6 & 120 & 120 &  600 \\
Samsung Galaxy S25 Ultra OS SxxR        &  5 & 100 &  5 & 100 & 100 &  400 \\
Samsung Galaxy S25 Ultra OS Sxx5        &  3 &  60 &  3 &  60 &  60 &  120 \\
Samsung Galaxy S25 Ultra OS SxxB        &  2 &  40 &  2 &  40 &  40 &   40 \\
Apple iPhone 17 Pro Max iOS 26.2    &  4 &  80 &  4 &  80 &  80 &  240 \\
Apple iPhone 17 Pro Max iOS 26.1    &  4 &  80 &  4 &  80 &  80 &  240 \\
Apple iPhone 17 Pro Max iOS 26.0.1  &  2 &  40 &  2 &  40 &  40 &   40 \\
\hline
Total                               & 30 & 600 & 30 & 600 & 600 & 1 920%
\end{tabular}
\caption{\myLabel{table}{Description of the dataset and experiment statistics}. The dataset consists of a total of 1 200 images acquired with 30 smartphones of 8 different software versions, among three brands. Mismatching correlations of rotated test images are not counted. For example, for the Google Pixel 10 Pro XL, 4 instances and 40 images per smartphone are collected; these images are split into 20 references images and 20 test images per instance, leading to a total of $4\times20=80$ reference images and $4\times20=80$ test images;  a reference PRNU is computed per instance, counting 4 in total;  for each of the 80 test images, 80 matching correlations can be computed, relative to the corresponding PRNU reference of the same smartphone; for each of the 80 test images, $3\times80=240$ mismatching correlations can be computed relative to the PRNU references of the 3 other smartphones of the same class.  For the Samsung software version, SxxR, Sxx5, SxxB correspond to 
S938BXXS7BYLR, S938USQSBCZF5, S938USQS8BZBB, respectively.}
\label{table:dataset}
\end{table*}

\begin{table}
\resizebox{\linewidth}{!}{
    \begin{tabular}{ c | c }
     Camera model & Count of metadata profiles\\\hline
     Google Pixel 10 Pro XL & 4 (1), 2 (3) and 1 (10)\\
     Google Pixel 10 Pro & 6 (1), 2 (1) and 1 (9)\\
     \hline
     Samsung Galaxy S25 Ultra & 5 (1), 3 (1), 2 (9) and 1 (41)\\
     \hline
     Apple iPhone 17 Pro Max & 4 (2), 3 (1), 2 (3) and 1 (42)\\
    \end{tabular}
}
    \caption{\myLabel{table}{Count of users per smartphone model metadata profile} with at least
    40 images, including 20
    portrait ones. $X (Y)$ means that there are $Y$ metadata profiles common to $X$ users.}
\end{table}

\subsection{\myLabel{section:\ImageSourceIdentification}{\sPCEbasedPRNUDetection}}

    To determine whether a query image $\textbf{I}$, with noise residual $\textbf{R}$, was taken by a given camera %
    with estimated PRNU $\hat{\textbf{K}}$, one popular method is to compute the signed Peak-to-Correlation Energy ($\text{sPCE}$) between $\textbf{R}$ and $\textbf{I}\hat{\textbf{K}}$, as introduced by~\cite{Goljan2008}:%
    \begin{equation}\label{eq:sPCE}
        \text{sPCE} (\textbf{R},\textbf{I}\hat{\textbf{K}}) = \frac{\operatorname{sign}\left(\textbf{C}_{\textbf{R},\textbf{I}\hat{\textbf{K}}}(0, 0)\right)\cdot\textbf{C}_{\textbf{R},\textbf{I}\hat{\textbf{K}}}^2(0, 0)}{\frac{1}{MN - |\textbf{A}|}\sum\limits_{(x, y) \notin \textbf{A}} \textbf{C}_{\textbf{R},\textbf{I}\hat{\textbf{K}}}^2(x, y)},
    \end{equation}
    where $M$ and $N$ denote the height and width of the images, $\textbf{A}$ is an exclusion square of size $5\times 5$ around $(0, 0)$, and $\textbf{C}_{\textbf{R},\textbf{I}\hat{\textbf{K}}}(x, y)$ is the unnormalized circular cross-correlation between $\textbf{R}$ and $\textbf{I}\hat{\textbf{K}}$, defined as:
    \begin{dmath}
        \textbf{C}_{\textbf{R},\textbf{I}\hat{\textbf{K}}}(x, y) = \frac{\sum_{(m, n)}\textbf{R}(m, n)\textbf{I}\hat{\textbf{K}}(m + x \mathbin{\scriptstyle\bmod} M, n + y \mathbin{\scriptstyle\bmod} N)}{MN}.
    \end{dmath}
     Intuitively, the sPCE normalizes the cross-correlation by the mean squared cross-correlation over spatial shifts, making it more robust to non-unique artifacts.
    
     In practice, source attribution is performed by comparing the resulting sPCE value against a decision threshold: given a positive decision threshold $t$ we decide that the estimated PRNU $\hat{\textbf{K}}$ is present in the image under inspection $\textbf{I}$—and therefore that $\textbf{I}$ originates from the corresponding camera—if and only if 
     \begin{equation}
         \text{sPCE}(\textbf{R}, \textbf{I}\hat{\textbf{K}}) > t.
    \end{equation}

    \section{\label{section:\ImageSourceIdentificationPRNUDetectionDifferenceOfsPCEsBasedPRNUDetectionLabel}Proposed metric: $\Delta\text{sPCE}$}%
    
    As we will show in Section~\myRef{section}{\Experiments}, the sPCE approach can have limitations, especially for modern smartphone cameras. For this reason, we introduce an alternative metric: the $\Delta\text{sPCE}$. For a given candidate camera and query image, $\Delta\text{sPCE}$ is defined as the difference between the sPCE obtained between the query image and the candidate camera's estimated PRNU and the mean sPCE obtained for the same image with the estimated PRNUs of the other cameras sharing the same model and software version. More precisely, given a query image $\textbf{I}$ with noise residual $\textbf{R}$, a candidate PRNU $\hat{\textbf{K}}$, and a collection of $K$ distinct cameras other than the one under analysis, with respective estimated PRNUs $\lbrace \hat{\textbf{K}}_{i} \rbrace _{i = 1, \dots, K}$,
    we compute:
        \begin{dmath}
        \Delta\text{sPCE}(\textbf{R}, \textbf{I}, \hat{\textbf{K}}, \lbrace \hat{\textbf{K}}_{i} \rbrace _{i = 1, \dots, K}) =  \text{sPCE}(\textbf{R},\textbf{I}\hat{\textbf{K}}) - \frac{1}{K}\sum_{i=1}^K\text{sPCE} (\textbf{R},\textbf{I}\hat{\textbf{K}}_{i}).
    \end{dmath}
    
    In practice, source attribution is performed by comparing the resulting $\Delta\text{sPCE}$ value against a decision threshold: given a positive decision threshold $t_\Delta$, we
    decide that the estimated PRNU $\hat{\textbf{K}}$ is present in the image under inspection $\textbf{I}$—and therefore that $\textbf{I}$ originates from the corresponding camera—if and only if 
    \begin{equation}
        \Delta\text{sPCE}(\textbf{R}, \textbf{I}, \hat{\textbf{K}}, \lbrace \hat{\textbf{K}}_{i} \rbrace _{i = 1, \dots, K}) > t_\Delta.
    \end{equation}

    Note that this condition 
    is equivalent to 
    \begin{equation}
    \text{sPCE}(\textbf{R},\textbf{I}\hat{\textbf{K}}) > t_\Delta + \frac{1}{K}\sum_{i=1}^K\text{sPCE} (\textbf{R},\textbf{I}\hat{\textbf{K}}_{i}).
    \end{equation}
Thus, rather than applying a fixed sPCE threshold to all images and candidate cameras, $\Delta\text{sPCE}$ effectively introduces an adaptive threshold that depends on both the query image and the candidate camera. The underlying intuition is that, when the candidate camera is the true source, its estimated PRNU should yield a higher sPCE with the query image than the estimated PRNUs of the other cameras. In that case, we indeed expect:
\begin{equation}
\text{sPCE}(\textbf{R},\textbf{I}\hat{\textbf{K}}) > \frac{1}{K}\sum_{i=1}^K\text{sPCE} (\textbf{R},\textbf{I}\hat{\textbf{K}}_{i}).
\end{equation}

To ensure a given operational false positive rate, $t_\Delta$ is a necessary safety margin from the sPCE mean of other cameras' PRNU estimates.%

\begin{figure*}[t]
    \includesvg[inkscapelatex=false,width=\textwidth]{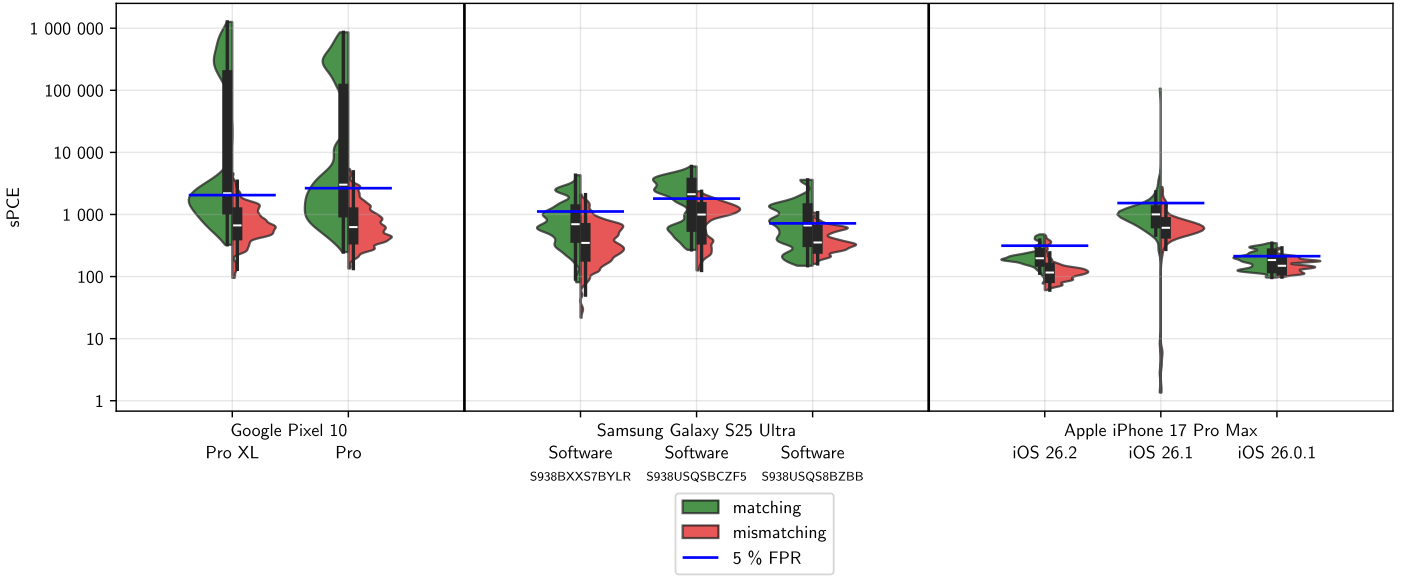}
    \caption{\label{fig:sPCE}\myLabel{figure}{sPCEs computed per smartphone-software-version}. sPCEs between a test image noise residual and an estimated PRNU of the same user are reported in \textbf{green} and sPCEs between a test image noise residual and an estimated PRNU of a different user are reported in \textbf{red}. The sPCE threshold per software version inferred from 5\% FPR are drawn in \textbf{blue}. While sPCEs between a test image and an estimated PRNU of the same user are a bit higher than sPCEs between a test image and estimated PRNUs of different users, the Apple iPhone 17 Pro Max show very similar distributions. As a result, setting a sPCE threshold to separate the best both distributions is ineffective.}
\end{figure*}

\begin{figure*}[t]
    \includesvg[inkscapelatex=false,width=\textwidth]{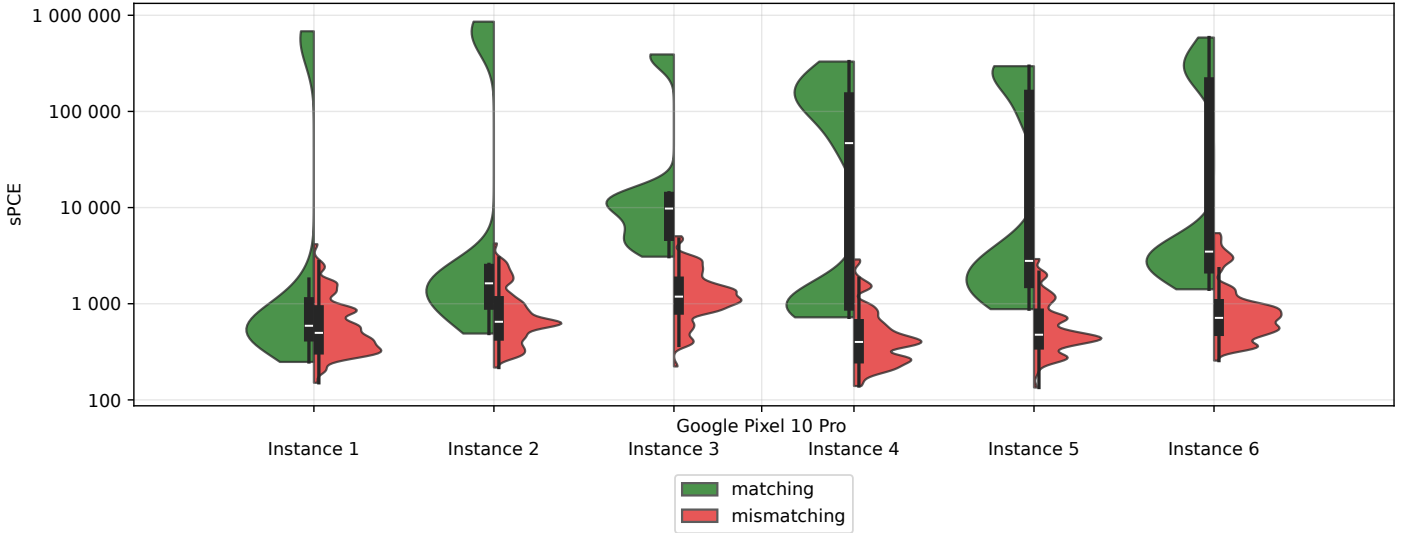}
    \caption{\myLabel{figure}{sPCEs computed for Google Pixel 10 Pro smartphones}. sPCEs between a test image noise residual and an estimated PRNU of the same user are reported in \textbf{green}; sPCEs between a test image noise residual and an estimated PRNU of a different user are reported in \textbf{red}. From one instance to another, there is a very similar distribution of sPCEs between test images and the estimated PRNUs of different users. However, the distribution of sPCEs between test images and the same user's estimated PRNU varies widely from one instance to another, even when accounting for the two sPCE ranges. This difference is especially visible in instance 3.}
\end{figure*}

\begin{figure*}[t]
    \includesvg[inkscapelatex=false,width=\textwidth]{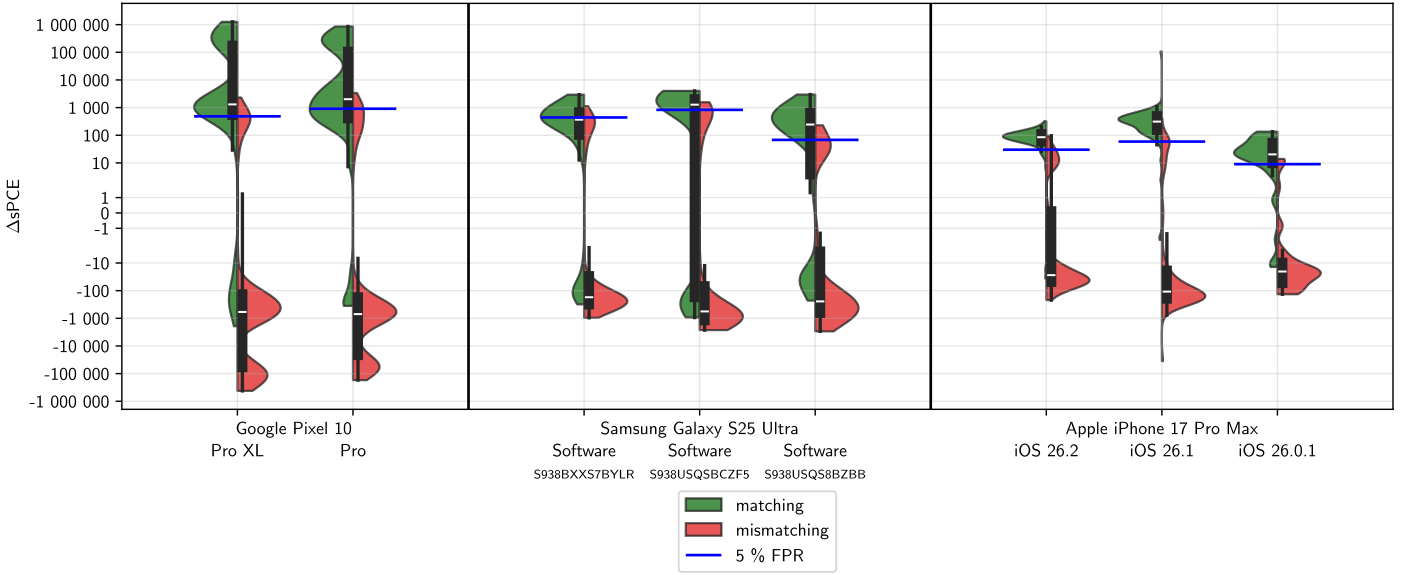}
    \caption{\label{figure:DeltasPCEs computed per software version from Google, Samsung and Apple smartphones}$\Delta$sPCEs computed per smartphone-software-version. $\Delta$sPCEs testing the correct source camera are denoted in \textbf{green}, while $\Delta$sPCEs testing an incorrect source camera are denoted in \textbf{red}. The $\Delta$sPCE threshold per software version inferred from 5\% FPR are drawn in \textbf{blue}. %
    Apple iPhone 17 Pro Max $\Delta$sPCEs testing the correct source camera are significantly higher than $\Delta$sPCEs testing an incorrect source camera. As presented in Table~\myRef{table}{Observed TPR}, this substantially improves the separability of both same source and different source distributions thanks to a $\Delta$sPCE threshold compared to a sPCE threshold shown in Fig.~\myRef{figure}{sPCEs computed for Google Pixel 10 Pro smartphones}.}
\end{figure*}

\begin{figure*}[t]
    \includesvg[inkscapelatex=false,width=\textwidth]{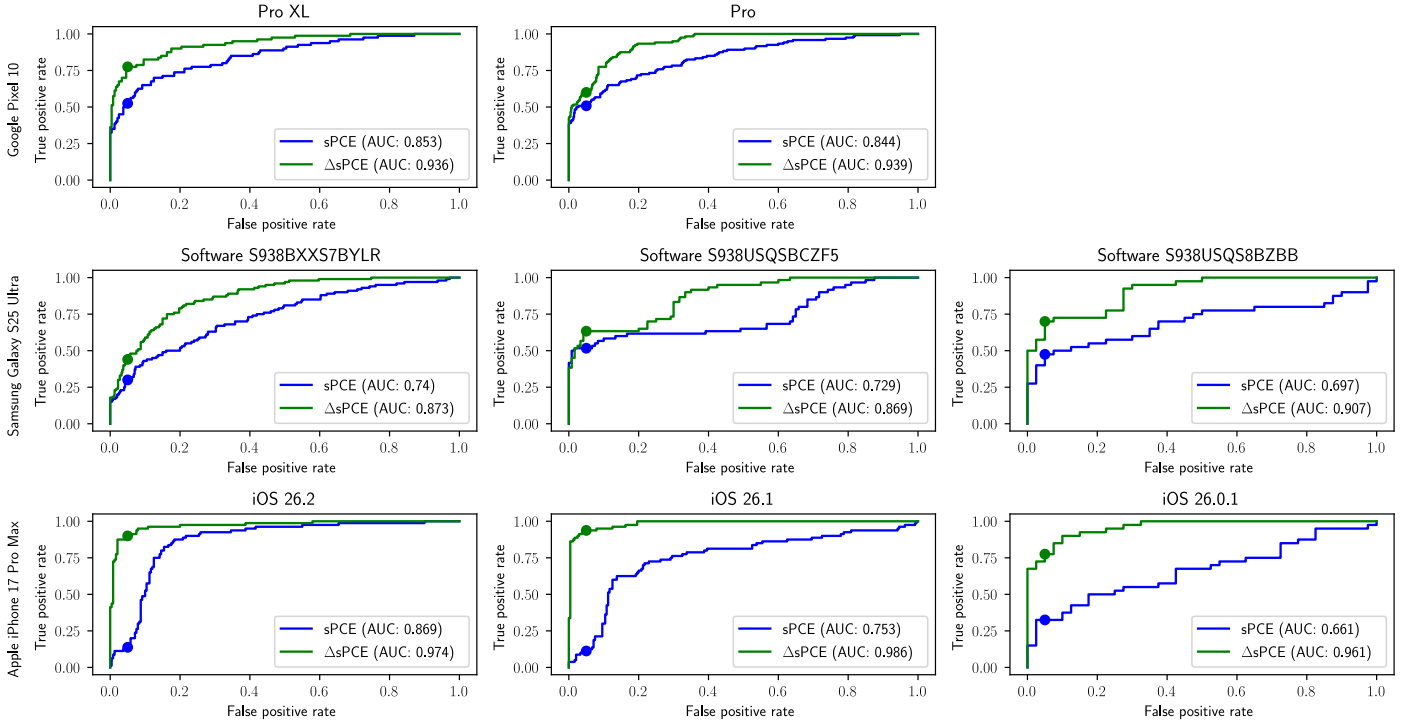}
    \caption{\myLabel{figure}{ROC curves computed from Google, Samsung and Apple smartphones}.
    The true positive rate with a 5\% false positive rate is denoted by a dot. Green ROC curves rely on a $\Delta$sPCE threshold and blue ROC curves rely on a sPCE threshold.%
    }
\end{figure*}

\mySection{Dataset}

\newcommand{\numberOfImagesByUserSet}{20\xspace}

\newcommand{\numberOfImagesByUser}{\number\numexpr 2 * \numberOfImagesByUserSet}
\newcommand{\numberOfUsersPerModel}{10\xspace}

To measure the impact of modern smartphone image processing pipeline on PRNU detection, we collected images from multiple smartphones of different popular recent models. In the following, we consider only images without digital zoom, that is, with Exif metadata \emph{Digital Zoom Ratio} 1 if the brand is not Google and 0 otherwise.
To consider for each smartphone model the most representative and similar fingerprint images, we focused on images having one of the most common metadata profiles across users. %
By the metadata profile of an image, we denote its resolution and its Exif tags. For most tags, only their presence or absence is relevant, disregarding the corresponding values; however, the value is relevant for the following tags:
\begin{enumerate}
    \item \emph{Make} and \emph{Model}, which must be present and match the considered smartphone brand and model;
    \item \emph{Focal Length} which must be present and \emph{Focal Length (35mm format)} for lens selection; %
    \item \emph{Software}, which must be present.
\end{enumerate}
Both the specified value for \emph{Digital Zoom Ratio} and identical \emph{Focal Length} values are important to estimate the maximum resolution PRNU estimation of a single camera among the multiple ones of each smartphone. %
Our assumption is that for a given smartphone model and software version, the most widespread metadata profile, designated as \emph{smartphone-software-version}, denotes its default processing.
While such default processing likely excludes unexpected correlation patterns likely involved in unusual shooting modes and brand-proprietary technologies, it also makes camera identification harder, as the images result from the same image processing pipeline. As a comparison, \cite{9393356} do not compare same smartphone-software-version noise residuals, but only enforce consistent mandatory \emph{Make} and \emph{Model}, and \emph{Focal Length} if available.
For each smartphone-software-version considered, we collect data from \numberOfUsersPerModel distinct users, with \numberOfImagesByUser images per user, including at least \numberOfImagesByUserSet portrait images. For each smartphone model, multiple software versions are considered to reach \numberOfUsersPerModel users. To this end, for the Google Pixel 10, the Pro variant is considered in addition to the Pro XL variant. 
For each smartphone-software-version, its \numberOfImagesByUser images are uniformly split into a reference set and a test set, with only portrait images in the reference set.

Table~\myRef{table}{Count of users per smartphone model metadata profile} shows the count of users per smartphone model metadata profile with at least \numberOfImagesByUser images, including \numberOfImagesByUserSet portrait ones. It highlights the existence of many metadata profiles, despite the mentioned constraints,
hence the camera assignment difficulty due to considered instances having very similar image processing pipelines.
As an order-of-magnitude estimate, this filter was applied to the mentioned Google, Samsung, and Apple models, on respectively, 10~261, 15~093, and 12~411 images. %
Table~\myRef{table}{Description of the dataset and experiment statistics} describes the dataset, consisting of a total of 1 200 images acquired with 30 smartphones of 8 different software versions, such that there are 10 different users for each smartphone model.

\mySection{\Experiments}

Each camera's PRNU is estimated from \numberOfImagesByUserSet portrait-oriented images to reduce orientation ambiguity, under the assumption that portrait images are rarely captured upside-down. For each test image, we compute sPCE for both its original and its 180° rotation, retaining the higher score. When the test image and reference fingerprints have different orientations, the test image is first rotated by 90°. %

Only the green color channel of each image is considered. The green color channel has a higher density of sensors than the other color channels, so interpolation artifacts due to demosaicking are minimized.
For each camera, the reference image set is used to estimate the camera PRNU using Eq.~\ref{eq:mle}.
Then, the sPCE (Eq.~\ref{eq:sPCE}) is computed for each image of the test set, relative to each reference PRNU sharing the same smartphone-software-version as the test image.

Table~\myRef{table}{Description of the dataset and experiment statistics} summarizes the experiment conditions. It emphasizes the very precise filter applied over 37~765 images taken from 30 instances, having 125 metadata profiles leading to 8 smartphone-software-versions, totalling 600 reference images and 600 test images, and 600 matching and 1 920 mismatching correlations.

As shown next, our experiments confirm the limitations of the sPCE approach for modern smartphone cameras, as first described by~\cite{9393356}. These experiments also show that $\Delta$sPCE, introduced in Section~\myRef{section}{\ImageSourceIdentification:\ImageSourceIdentificationPRNUDetection:\ImageSourceIdentificationPRNUDetectionDifferenceOfsPCEsBasedPRNUDetection}, mitigates these limitations.

\newcommand{\ExperimentssPCEBasedSourceCameraAttribution}{sPCE-based source camera attribution}
\subsection{\label{section:\Experiments:\ExperimentssPCEBasedSourceCameraAttribution}\ExperimentssPCEBasedSourceCameraAttribution}

Fig.~\ref{fig:sPCE} 
shows the distributions of matching and mismatching sPCE scores for the different smartphone models and software versions. While matching scores are mostly higher than mismatching scores, there is a substantial overlap in all cases, especially for the iPhone 17 Pro Max.
Consequently, using a fixed sPCE threshold such as 60 in~\citep{largescale_goljan} does not enable distinguishing matching from mismatching comparisons.
The score distributions also vary across software versions of the same smartphone model, most notably between iOS 26.1 and the other iOS versions.
These differences may result from changes in the image processing pipeline or acquisition conditions, including the scenes captured.

For the Google Pixel 10 Pro XL and non-XL, the matching sPCE distributions exhibit two distinct score ranges. All test images yielding exceptionally high scores are landscape-oriented, although the reference PRNUs were only estimated from portrait images. However, not all landscape images produce high scores. The origin of this behaviour remains unclear. Figure~\myRef{figure}{sPCEs computed for Google Pixel 10 Pro smartphones} details the score distributions for individual Google Pixel 10 Pro instances. While mismatching scores follow similar distributions across instances, matching scores vary considerably, even when accounting for the two score ranges.

Source camera attribution performance is evaluated using the ROC curves in Fig.~\myRef{figure}{ROC curves computed from Google, Samsung and Apple smartphones} and the true positive rates (TPRs) reported in Table~\myRef{table}{Observed TPR}.
We consider a false positive rate (FPR) of 5\%, which is too high for operational forensic applications but provides a common evaluation point given the limited dataset size. In particular, smartphone groups in the dataset contain as few as 40 mismatching comparisons (see Table~\ref{table:dataset}), which prevents a reliable evaluation at substantially lower FPRs. At 5\% FPR, conventional sPCE performs particularly poorly on the iPhone 17 Pro Max running iOS 26.1, achieving a TPR of only 11.2\%.

\newcommand{\ExperimentsDifferencesOfsPCEBasedSourceCameraAttribution}{Difference of sPCEs-based source camera attribution}
\subsection{\label{section:\Experiments:\ExperimentsDifferencesOfsPCEBasedSourceCameraAttribution}$\Delta$sPCE-based source camera attribution}

Fig.~\myRef{figure}{DeltasPCEs computed per software version from Google, Samsung and Apple smartphones} shows the distribution of matching and mismatching $\Delta$sPCE scores for each smartphone model and software version. Compared with the sPCE distributions in Fig.~\myRef{figure}{sPCEs computed per smartphone-software-version}, $\Delta$sPCE provides substantially better separation for all smartphone models, especially for the iPhone 17 Pro Max. The predominantly negative $\Delta$sPCE scores for mismatching comparisons can be explained by the presence of the true source camera among the background candidates, whose sPCE is generally higher than those of incorrect candidates. As a result, red and green $\Delta$sPCE distributions look vertically symmetric. %
One can see a substantial improvement in the separability of the matching (green) and mismatching (red) distributions relative to Fig.~\ref{fig:sPCE}. All smartphone-software-version cases show a good separability, while the most marked difference relative to Fig.~\ref{fig:sPCE} is for the case of the
Apple iPhone 17 Pro Max. %
The improvement of separability of both distributions with a $\Delta$sPCE threshold is less significant for other smartphone models than the Apple iPhone 17 Pro Max.

Fig.~\myRef{figure}{ROC curves computed from Google, Samsung and Apple smartphones} compares the ROC curves obtained using sPCE and $\Delta$sPCE. The AUC improves for all smartphone models and software versions, with particularly substantial gains for the iPhone 17 Pro Max. For iOS 26.0.1, the AUC increases from 0.661 to 0.961, and for iOS 26.1, from 0.753 to 0.986. However, since forensic applications require low false positive rates, the true positive rate at a given (low) FPR is more informative than the overall AUC.

Table~\myRef{table}{Observed TPR} reports the true positive rates (TPRs) at 5\% FPR. %
$\Delta$sPCE outperforms conventional sPCE across all smartphone models and software versions, with particularly large improvements for the iPhone 17 Pro Max. For iOS 26.1, the TPR increases from 11.2\% to 93.8\%, while for iOS 26.2, it increases from 13.8\% to 90\%. These results demonstrate the benefits of query-dependent score normalization under the considered experimental conditions.

\begin{table*}
    \centering
    \begin{tabular}{ c || c | c || c | c || c ||}
      & \multicolumn{2}{c||}{sPCE} & \multicolumn{2}{|c||}{$\Delta$sPCE} & \\
     Smartphone software version & TPR & $t$ & TPR & $t_\Delta$ & TPR difference\\
     \hline
     Google Pixel 10 Pro XL & 52.5 & 2~051 & \textbf{77.5} & 483 & 25.0\\
     Google Pixel 10 Pro & 50.8 & 2~652 & \textbf{60.0} & 908 & 9.2\\
     \hline
     Samsung Galaxy S25 Ultra OS S938BXXS7BYLR & 30.0 & 1~117 & \textbf{44.0} & 440 & 14.0\\
     Samsung Galaxy S25 Ultra OS S938USQSBCZF5 & 51.7 & 1~802 & \textbf{63.3} & 828 & 11.6\\
     Samsung Galaxy S25 Ultra OS S938USQS8BZBB & 47.5 & 719 & \textbf{70.0} & 68 & 22.5\\
     \hline
     Apple iPhone 17 Pro Max iOS 26.2 & 13.8 & 314 & \textbf{90.0} & 30 & 76.2\\
     Apple iPhone 17 Pro Max iOS 26.1 & 11.2 & 1~527 & \textbf{93.8} & 59 & 82.6\\
     Apple iPhone 17 Pro Max iOS 26.0.1 & 32.5 & 213 & \textbf{77.5} & 9 & 45.0\\
     \hline
    \end{tabular}
    \caption{\myLabel{table}{Observed TPR} (in \%) for 5\% FPR. When considering a $\Delta$sPCE threshold instead of a sPCE one, all software versions of all models have a significantly higher TPR.
    For instance, the Apple iPhone 17 Pro Max iOS 26.1 TPR increases from 11.2\% to 93.8\% by leveraging the $\Delta$sPCE threshold. $\Delta$sPCE TPR - sPCE TPR is denoted as \emph{TPR difference} measured in \%..}
\end{table*}

\mySection{Discussion}

\cite{9393356} showed that PRNU estimates from modern smartphones contain a component shared across devices of the same model, a consequence of increasingly complex image processing pipelines. This shared component inflates the sPCE between an image and the fingerprints of other devices with the same model and software version, which explains the high false-positive rates reported in the literature (Section~\myRef{section}{\RelatedWorks}).

$\Delta$sPCE addresses this inflation during scoring rather than during fingerprint estimation. It uses the fingerprints of the other candidate devices with the same model and software version as a background model. For each questioned image, it estimates a background score from other devices sharing the same processing pipeline, and subtracts it from the candidate camera's sPCE.
This is analogous to test normalization (T-norm), long used in speaker verification~\citep{10.1006/dspr.1999.0360}. T-norm additionally divides the score by the standard deviation of the background scores. We tested this variant, but it did not improve our results. A likely reason is that, with at most six background devices per image, the standard deviation estimate is too noisy.
The accuracy gains reported in Section~\myRef{section:\Experiments}{\ExperimentsDifferencesOfsPCEBasedSourceCameraAttribution} indicate that the device-specific PRNU signal survives modern processing pipelines: the shared component masks it but does not erase it. PRNU-based attribution among identical smartphones, therefore, remains a relevant goal.

Applying $\Delta\text{sPCE}$ in forensic practice would require calibrating its decision threshold using representative images from the same smartphone model and software version. Such calibration should ideally rely on camera instances independent of those used for evaluation.

Two limitations apply to these results. First, our dataset contains at most six devices per model-software version pair, which limits how firmly we can draw conclusions. Second, $\Delta$sPCE, as defined, depends on the set of candidate cameras: it assumes the true device is among them, and its value changes when that set changes. It is therefore not yet suitable for operational use, which generally requires one-to-one verification with a stable decision threshold.

Several questions remain before $\Delta$sPCE can become a practical tool:
\begin{itemize}

\item Identifying the model first. The background model requires knowing the model and software version of the questioned image. In an unknown setting, can these be determined reliably enough to select the right background, either from metadata or from model-level traces?

\item Size of the background. How many devices of the same model are needed to obtain a stable estimate of the shared component? In particular, does dividing by the standard deviation, as in the full T-norm, become effective with larger backgrounds? If so, it would put scores on a common scale across models and images, potentially leading to more universal decision thresholds.

\item Generalization. Does a background model built from default-mode images also work for non-standard processing, such as night, portrait, or zoomed modes, which have different metadata profiles?

\item Better background models. Would a fixed reference set of devices known not to be the suspect make the score independent of the candidate set? Would removing the shared component directly from the fingerprints be more accurate?

\end{itemize}

\mySection{Conclusion}

In this work, we investigated the reliability of PRNU-based source identification on recent smartphones from major brands. First, we showed that sPCE, with the threshold commonly used in the literature, does not account for non-unique artifacts (NUAs). As a result, sPCE-based attribution remains fragile even with default processing, confirming the results of~\cite{9393356}. Second, we showed that the proposed $\Delta$sPCE substantially improves attribution accuracy by accounting for the signal shared by devices with the same model and processing pipeline. This conclusion is preliminary, given the limited size of our dataset, and $\Delta$sPCE is not yet suitable for real-world cases in its current form. Nevertheless, our findings revive hope that PRNU-based smartphone attribution may be feasible even among devices that share the same hardware and software.

\mySection{Acknowledgements}
This project was provided with computing HPC and storage resources by GENCI at IDRIS thanks to the grant R1001016831 on the supercomputer Dalia.
This work benefited from Hi! PARIS and State funding managed by the French National Research Agency (ANR) under the France 2030 program, reference ANR-23-IACL-0005.

The research that originated the results presented in this publication was partly supported by the Programa de Desarrollo de Ciencias Básicas (PEDECIBA) of Uruguay.

Centre Borelli is also a member of Université Paris Cité, SSA and INSERM.

\bibliographystyle{cas-model2-names}

\bibliography{biblio}

\end{document}